**Digital Twin Technology Enabled Proactive Safety Application for Vulnerable Road Users: A Real-World Case Study**

**Erik Rúa\***
PhD Student
CINTECX – GeoTech Group
University of Vigo
Vigo University Campus, As Lagoas, Marcosende, Vigo, 36310, Spain
Email: erik.rua@uvigo.gal

**Kazi Hasan Shakib**
Ph.D. Student
Department of Civil, Construction & Environmental Engineering
The University of Alabama
3014 Cyber Hall, Box 870205, 248 Kirkbride Lane, Tuscaloosa, AL 35487
Email: khshakib@crimson.ua.edu

**Sagar Dasgupta**
Ph.D. Student
Department of Civil, Construction & Environmental Engineering
The University of Alabama
3014 Cyber Hall, Box 870205, 248 Kirkbride Lane, Tuscaloosa, AL 35487
Email: sdasgupta@crimson.ua.edu

**Mizanur Rahman, Ph.D.**
Assistant Professor
Department of Civil, Construction & Environmental Engineering
The University of Alabama
3015 Cyber Hall, Box 870205, 248 Kirkbride Lane, Tuscaloosa, AL 35487
Email: mizan.rahman@ua.edu

**Steven Jones, Ph.D.**
Director, Transportation Policy Research Center
The University of Alabama
3024 Cyber Hall, Box 870205, 248 Kirkbride Lane, Tuscaloosa, AL 35487
Email: sjones@eng.ua.edu

\*Corresponding author

Word count: 6,190 words text + 3 tables x 250 words (each) = 6,940words

Submission date: August 1, 2023

**ABSTRACT**

While measures, such as traffic calming and advance driver assistance systems, can improve safety for Vulnerable Road Users (VRUs), their effectiveness ultimately relies on the responsible behavior of drivers and pedestrians who must adhere to traffic rules or take appropriate actions. However, these measures offer no solution in scenarios where a collision becomes imminent, leaving no time for warning or corrective actions. Recently, connected vehicle technology has introduced warning services that can alert drivers and VRUs about potential collisions. Nevertheless, there is still a significant gap in the system's ability to predict collisions in advance. The objective of this study is to utilize Digital Twin (DT) technology to enable a proactive safety alert system for VRUs. A pedestrian-vehicle trajectory prediction model has been developed using the Encoder-Decoder Long Short-Term Memory (LSTM) architecture to predict future trajectories of pedestrians and vehicles. Subsequently, parallel evaluation of all potential future safety-critical scenarios is carried out. Three Encoder-Decoder LSTM models, namely pedestrian-LSTM, vehicle-through-LSTM, and vehicle-left-turn-LSTM, are trained and validated using field-collected data, achieving corresponding root mean square errors (RMSE) of 0.049, 1.175, and 0.355 meters, respectively. A real-world case study has been conducted where a pedestrian crosses a road, and vehicles have the option to proceed through or left-turn, to evaluate the efficacy of DT-enabled proactive safety alert systems. Experimental results confirm that DT-enabled safety alert systems were succesfully able to detect potential crashes and proactively generate safety alerts to reduce potential crash risk.

**INTRODUCTION**

Roadway safety is a global concern, with an estimated 1.3 million fatalities and 20 to 50 million injuries occurring each year in road traffic crashes (*1*). Based on statistics of 2021 in the United States, there were 42,915 reported deaths on the roads, with approximately 20% of them being Vulnerable Road Users (VRUs), which include pedestrians, bicyclists, other cyclists, and individuals using personal conveyances, as well as road workers on foot (*2*). Specifically considering pedestrians within the VRU category, which encompasses individuals on foot engaging in activities, such as walking, running, jogging, hiking, sitting, or lying down, there were 7,388 fatalities and an estimated 60,577 pedestrians were injured in traffic crashes (*3*). Pedestrian deaths accounted for 17% of all traffic fatalities and 2% of all individuals injured in traffic crashes in 2021 (*3*). In terms of location, a higher percentage of pedestrian fatalities occurred in urban areas (84%) compared to rural areas (16%) (*3*). Furthermore, 16% of pedestrian fatalities took place at intersections, while 75% at non-intersection locations. The remaining 9% of pedestrian fatalities happened in other areas (*3*).

Prior studies focus on either reactive approach or probalistic safety alerts generation approach using real-time data (*4*). Research on pedestrian road safety has shifted towards Intelligent Transportation Systems (ITS) (*5*), which utilize information and communication technologies, to proactive safety alerts. A proactive approach that aims to anticipate and prevent issues before they occur. Recently, connected vehicle data has been used to generate safety alerts in real-time (*6*). However, safety alert generation is reactive in nature as it does not evaluate all possible future safety critical scnerions holistically. Among technologies within proactive approach, Digital Twins (DT) is a promising technology that is expected to transform ITS-based applications from reactive to proactive safety and mobility applications (*7*, *8*).

Transportation Digital Twin (TDT) is a real-time digital representation of a physical transportation asset or process to improve the safety and mobility (*9*). It is essential to emphasize that a requirement for a DT representation of an object is the real-time synchronization between the virtual and physical entities through frequent updates of state information. DT systems are often compared to the Internet of Things (IoT) or Cyber-Physical Systems (CPS). However, unlike IoT and CPS systems, DT systems uniquely maintain a virtual twin of the physical asset. This virtual twin serves as a direct representation of the physical object and enables the performance of predictive analytics that would otherwise be impossible on the physical asset alone. Before the emergence of DT systems, multi-agent simulation (MAS) was a prevalent digital simulation approach for modeling systems. Yet, DT systems differ fundamentally from MAS implementations because they rely on real-time or near real-time data from physical sensors for their models and simulations.

Prior studies have addressed safety and mobility in transportation using DT approach either leveraging the DT concept partially or fully (*10–13*). Currently, prototype systems are under development, focusing on particular use cases to fully exploit the advantages of a TDT system. These applications include cooperative ramp merging (*14–16*), cooperative driving (*17*), and driver behavior modeling (*18*), which effectively demonstrate how TDT systems enhance the safety and efficiency of established transportation systems. Recently, authors in (*19*) study developed a DT framework for pedestrian safety in hard-in-the-loop simulation environment. However, to the best of the authors knowledge, no previous study has specifically DT based proactive VRU safety application using predictive parallel scenarios of potential safety critical scenarios using real-time states update of vehicles and VRUs. Thus, the objective of this study is to develop a DT-based framework to indentify future potential crash scenarios and generate proactive safety alerts for both pedestrians and vehicles, indicating the possibility of a potential crash between them.

To achieve this goal, the physical world elements, including pedestrians and vehicles, will be monitored using GPS, accelerometers, and gyroscopes to create their digital shadows. Then, parallel evaluations using trajectory prediction models of all possible interaction scenarios are developed to identify safety critical scenario of a collision between them. Safety alerts will be generated for both pedestrians and vehicles when a safety critical scenario is identified, following the safety principles of Advanced Driver-Assistance Systems (ADAS). A real-world case study is conducted by developing a DT-based framework for proactive VRU safety to evaluate the efficacy of a DT-based VRU safety application. Thus, this study has the following two primary contributions: (i) developing a DT-enabled framework for a proactive VRU safety alert system; (ii) conducting field experiments to evaluate and validate the efficacy of a DT-enabled VRU safety application.

**RELATED WORK**

In this section, relevant existing literature on Transportation Digital Twin (TDT), ranging from a broader perspective of safety and mobility in transportation to more specific pedestrian safety studies that align with our objective, are reviewed. Special emphasis has been given on analyzing whether the existing DT-based approaches have been applied in real-world case studies. Furthermore, the inclusion of prediction models has been thoroughly investigated, as it plays a critical role in conducting parallel evalautions of all possible safety critical scenarios proactively, which constitutes the core of the digital sibling component within the digital twin framework (*9*). Cases where predictions have not been employed can restrict their proactive capabilities, as the ability to anticipate future events is diminished, thereby making it challenging to take preemptive actions. A summary of existing DT-based studies is present in the **TABLE 1**. (*10*) developed A Mobility Digital Twin (MDT) framework that consisting of the DT for Human, Vehicle, and Traffic with their respective predictions. (*11*) proposed a DT-based adaptive traffic signal control (ATSC) systems framework, where two algorithms are introduced to improve intersection performance trough parallel simulation of traffic demand prediction.

As shown in **TABLE 1**, a considerable number of studies are focused on the vehicle-based DT. (*15*) and (*14*) developed DT paradigm for connected vehicles using ADAS that leverages vehicle-to-cloud (V2C) communication. They applied a Nonlinear Autoregressive Neural Networks (NARNNs) to predict the difference between the actual speed of the driver and the advisory speed calculated. They demonstrated their method for a cooperative ramp merging scenario through a real-world evalaution. (*12*) proposed a sensor fusion technology that combines camera information and cloud-based DT data to predict lane change behavior in other vehicles to ADAS application. (*13*) developed a DT simulation using the Unity game engine for CAVs. (*17*) presented DT for cooperative driving system in non-signalized intersections allowing connected vehicles to cross without full stops. They proposed an algorithm for the prediction of motion estimation.

The studies addressing DT for humans can be categorized into two distinct groups: driver-focused and pedestrian-focused. Regarding the first approach, (*18*) proposed a Digital Behavioral Twin to share driver behavioral models among connected vehicles in order to predict collisions with neighboring vehicles and enhance road safety. (*16*) developed a Driver Digital Twin (DDT) with online prediction capabilities to enable personalized lane change behavior and evaluated in a real field implementation scenario involving on/off-ramp mandatory lane changing. As for the second pedestrian-focused approach, (*19*) presented a DT framework that simulates interactions between connected vehicles and pedestrians, utilizing individual DT for each entity and calculating the time-to-collision (TTC) between them. The works proposed by (*6*) and (*20*) do not specifically focus on the digital twin approach; however, their methos are relevant to this study as they address key aspects necessary for DT development. (*6*), presented a vision-based approach utilizing traffic cameras to generate real-time collision warnings based on average time-to-collision calculations, with a real-world validation performance by developing a connected vehicle based pedestrian safety application. On the other hand, (*20*) presented a method for vehicle-pedestrian path prediction and collision risk estimation that incorporates driver and pedestrian awareness which was validated with real-world data with 93 vehicle-pedestrian encounters.

The development of a Digital Twin (DT) for vehicles and pedestrians to predict and generate safety alerts, considering all possible future safety-critical scenarios in a real-world context, is an area lacking in current literature. The development of this DT for trajectory prediction through parallel evaluation of potential safety-critical scenarios enables the identification of the most safety-critical scenario, i.e., a crash risk between a vehicle and a pedestrian, proactively. Therefore, this paper focuses on the development of a real-world DT framework for VRU safety to identify safety-critical scenarios proactively and generate safety alerts as proactive measures.

**TABLE 1 Comparison between Existing TDT-based VRU Safety related Studies and Our Study.**

| Study | Focus | Predictive Analytics | Real-world scenario | Approach |
|---|---|---|---|---|
| (10) | Vehicle Human Traffic | Prediction for human, vehicle, and traffic | No | Developed A Mobility Digital Twin (MDT) |
| (11) | Adaptive traffic signal control (ATSC) systems | Parallel simulation of traffic demand prediction | No | DT-based ATSC systems framework to improve intersection performance |
| (15) (14) | Vehicle | Speed tracking error | Cooperative ramp merging | DT for connected vehicles using ADAS |
| (12) | Vehicle | No | No | Sensor fusion technology integrating cloud-based Digital Twin data to predict lane change behavior |
| (13) | Vehicle | No | No | DT for cavs using Unity |
| (17) | Vehicle | Motion estimation | No | DT for cooperative driving system in non-signalized intersections |
| (18) | Driver | Collision among neighboring vehicles | No | Digital Behavioral Twin to share driver models among connected vehicles |
| (16) | Driver | Lane change maneuver | On/off-ramp mandatory lane changing | Driver Digital Twin to enable personalized lane change behavior |
| (19) | Pedestrian Vehicle | Time to Collision | No | DT framework for interactions between connected vehicles and pedestrians |
| (6) | Pedestrian Vehicle | Time to Collision | Pedestrian in signalized crosswalk warning (PSCW) | Vision-based approach to generate real-time collision warnings |
| (20) | Pedestrian Vehicle | Path prediction | 93 vehicle-pedestrian encounters | Method for vehicle-pedestrian path prediction and collision risk estimation |
| **Our proposed work** | Pedestrian Vehicle | Parallel evaluations of vehicle and pedestrian path prediction | Scenarios in non-intersection zones of urban areas | DT for pedestrian and vehicle to identify the worst-case scenario |

**FRAMEWORK FOR DT-BASED PROACTIVE VRU SAFETY**
This section presents a conceptual and a real-world implementation framework of a DT-enabled proactive VRU safety alert system.

**Concepptual Framework**
        **FIGURE** 1 illustrates a DT-enabled proactive VRU safety framework. This approach builds upon the genralized reference architecture originally presented by our resarch group (*9*), with appropriate adaptations made to suit our specific application requirements. The framework consists of three key components: physical world, communication gateway and digital twin. The physical world of the presented framework consists of pedestrians and vehicles that are assumed to be connected in real time with the digital twin via the communication gateway. Pedestrians and vehicles are equipped with position sensors –i.e., Global Navigation Satellite Systems (GNSS) and inertial sensors (speedometers, accelerometers, gyroscopes)– are connected to the digital twins to capture their real-time state updates due to their dynamic behavior. Although the nature of the road network is static, the information available at state and federal transportation agencies is placed and adapted regularly on the digital twin server. The communication gateway serves to facilitate the communication between the physical world layer and the digital twin layer. The communication of the presented framework is physical-to-virtual (P2V) bidirectional, since the virtual part receives the data from the sensors of the physical world, and the agents of the physical world (pedestrians and vehicles) receive the proactive safety alerts.
        The digital twin layer is composed of three sub-layers: digital shadow, digital sibling, and application. Digital shadow acts as a digital replica of the physical elements encompassed within the physical world layer. Data from pedestrian and vehicle sensors is processed to infer missing information, aggregate data, fuse data and synchronize it before storing it in a database. Through the utilization of map matching techniques, the sensor data can be effectively integrated with the predefined paths of the road network. This integration enables the visualization of pedestrian and vehicle movements within the road network, providing a comprehensive understanding of their trajectories and interactions. Although the data is processed in the digital shadow layer, the database in which both historical and real-time data are stored belongs to the digital sibling layer. With the historical data from the database, an Encoder-Decoder LSTM model is trained to obtain a trained model to predict with the real-time data the distance that a pedestrian or a vehicle will travel in their respective paths. The predicted distances will be used to evaluate parallely all possible scenarios at the intersection (right-turn, through, and left-turn), determining through an algorithm whether a vehicle-pedestrian crash could occur. The application of the presented DT is the generation of proactive safety alerts of a potential crash between a vehicle and a pedestrian. These messages notify proactively that a possible crach may occur, so that if either of the two agents reacts by modifying its trajectory in advance, the possible collision can be avoided.

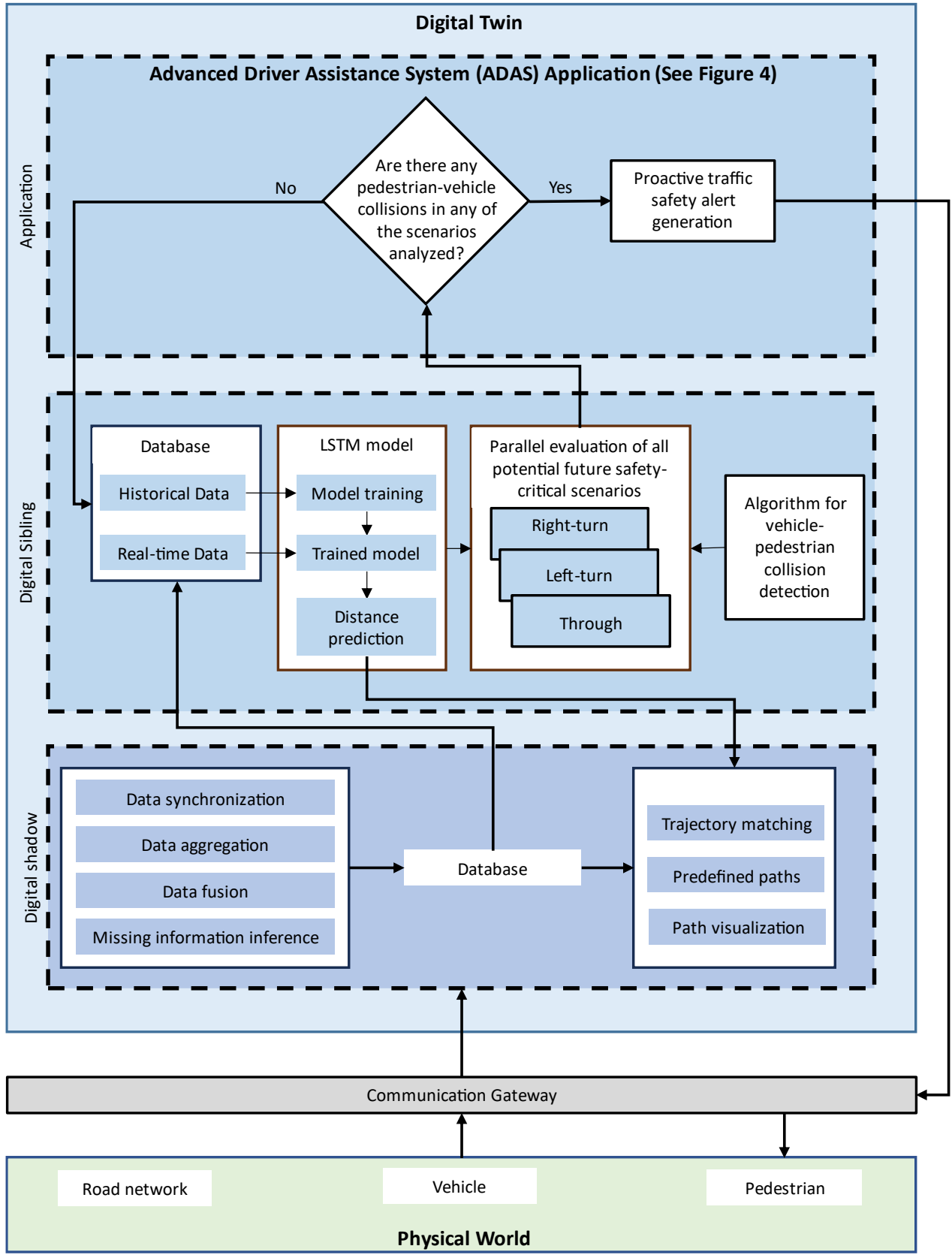

**FIGURE 1 DT-based proactive VRU safety framework.**

**Real-world Implementaiton Framework**

Following the three layers of the presented framework in the previous section, a implementation framework is presented, which will be evaluated and validated through a real-world case study.

*Physical world*

The implementation of the DT framework is conducted on Peter Bryce Boulevard, located within the campus of the University of Alabama in Tuscaloosa (Alabama, USA). Vehicles travel north-west bound limited to 25 mph without encountering any stop signs before reaching the crosswalk. The roadway has three lanes: the bus lane, the through lane, and the left-turn lane. For the purposes of this study, we focus on the possible crash between vehicles and pedestrians when driving through or left-turn, ignoring the bus lane. Two detection zones are defined: one for pedestrians and another for vehicles as shown in **FIGURE 2**. These zones are used to determine whether there are any pedestrians or vehicles to predict potential crashes. If no pedestrians are detected within the pedestrian detection zone, the digital twin will not be initiated as shown in **FIGURE 2**. Similarly, if pedestrians are detected but no vehicles are present within the vehicle detection zone, the digital twin will not be activated either. The pedestrian detection zone is defined by the width and length of the crosswalk. On the other hand, the vehicle detection zone is determined by a start distance from the crosswalk along the roadway and a final distance from the crosswalk, which form the boundaries of the vehicle detection zone.

Final distance is determined by the Stopping Sight Distance (SSD), which is the minimum sight distance required for a driver to stop without colliding, plus the distance covered by the vehicle in one second. The reason for adding the distance covered by the vehicle in one second is due to the frequency of data from the vehicle's sensors, which is 1 Hz. If the vehicle is detected at a distance corresponding to the SSD, receiving the alert message from the digital twin would be too late to avoid a collision. Therefore, the Final distance would be 42.24 meters of SSD (calculated at 25 mph (21)) plus 11.18 meters of the distance covered in 1 second (also calculated at 25 mph), resulting in a total of 58.42 meters.

Start distance is defined by the distance covered by the vehicle during the time it takes for the pedestrian to cross the entire crosswalk. This ensures that if the pedestrian starts crossing at one end and no vehicle reaches the crosswalk until the pedestrian reaches the other end, there is no risk of collision. Start distance calculated for 25 mph is 167.64 meters. **FIGURE 2** depicts these two distances in conjunction with the rest of the described scenario.

Regarding the experimental setup, both the pedestrian and the vehicle were equipped with specific sensors to facilitate data collection. Each participant carried an Inertial Measurement Unit (IMU) capable of measuring accelerations and rotations, along with a Global Navigation Satellite System (GNSS) sensor for positional tracking. These sensors were connected to a laptop, which served as the data acquisition unit. For the pedestrian, all the sensors were integrated into a backpack. The used GNSS sensor for the pedestrian was the Inertial Sense EVB-2. On the other hand, for the vehicle, a compact enclosure named CPT7700 was utilized for the vehicle. This enclosure contained an OEM7700 GNSS receiver and a high-performing Honeywell HG4930 Micro Electromechanical System (MEMS) Inertial Measurement Unit (IMU). **FIGURE 3-A** shows the installed computing unit on the roadside during the case study, **FIGURE 3-B** shows the sensors installed in the vehicle and **FIGURE 3-C** shows the pedestrian crossing the crosswalk with sensors installed in their backpack.

The collected data for both the pedestrian and the vehicle serves a dual purpose: first, to train and validate the Encoder-Decoder LSTM model, and second, to conduct a comprehensive evaluation of the digital twin. For the pedestrian, data was collected at a frequency of 5 Hz, resulting in a total of 1,693 data points. In the case of the vehicle, data was collected at a frequency of 1 Hz, yielding a total of 643 data points for both scenarios, i.e., when the vehicle goes through and when it makes a left-turn. The following variables were recorded for all cases: timestamp, longitude, latitude, acceleration, velocity, and angular velocity (for all three axes).

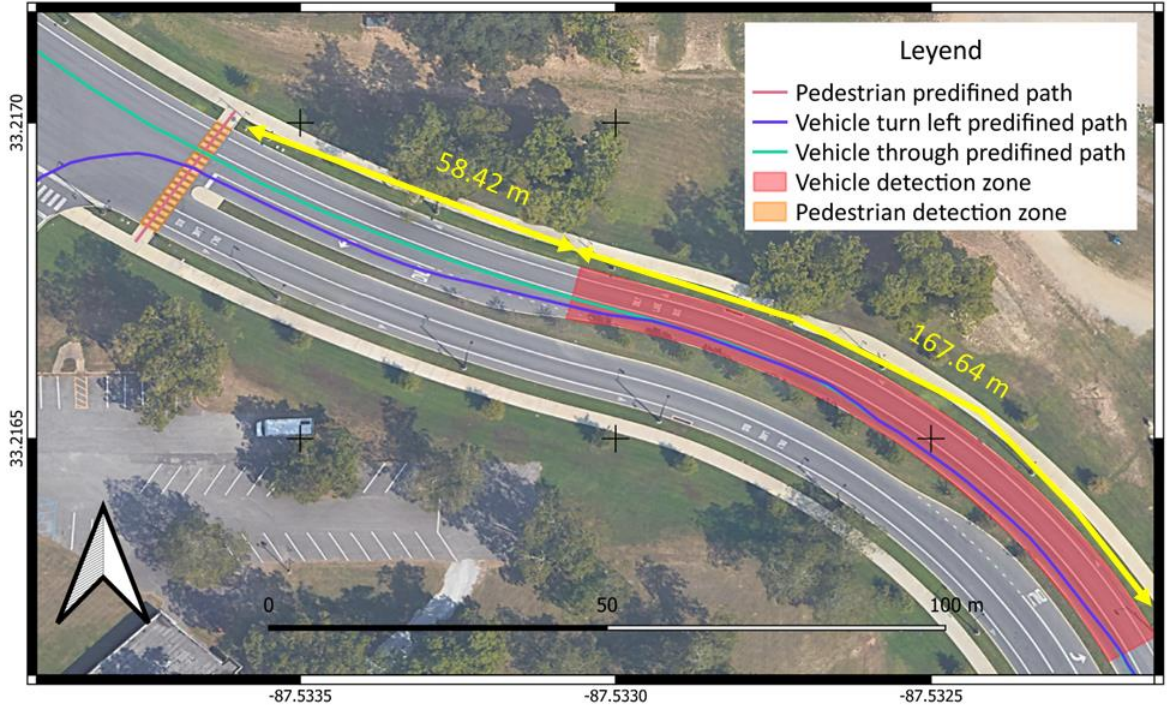

**FIGURE 2 Case study area**

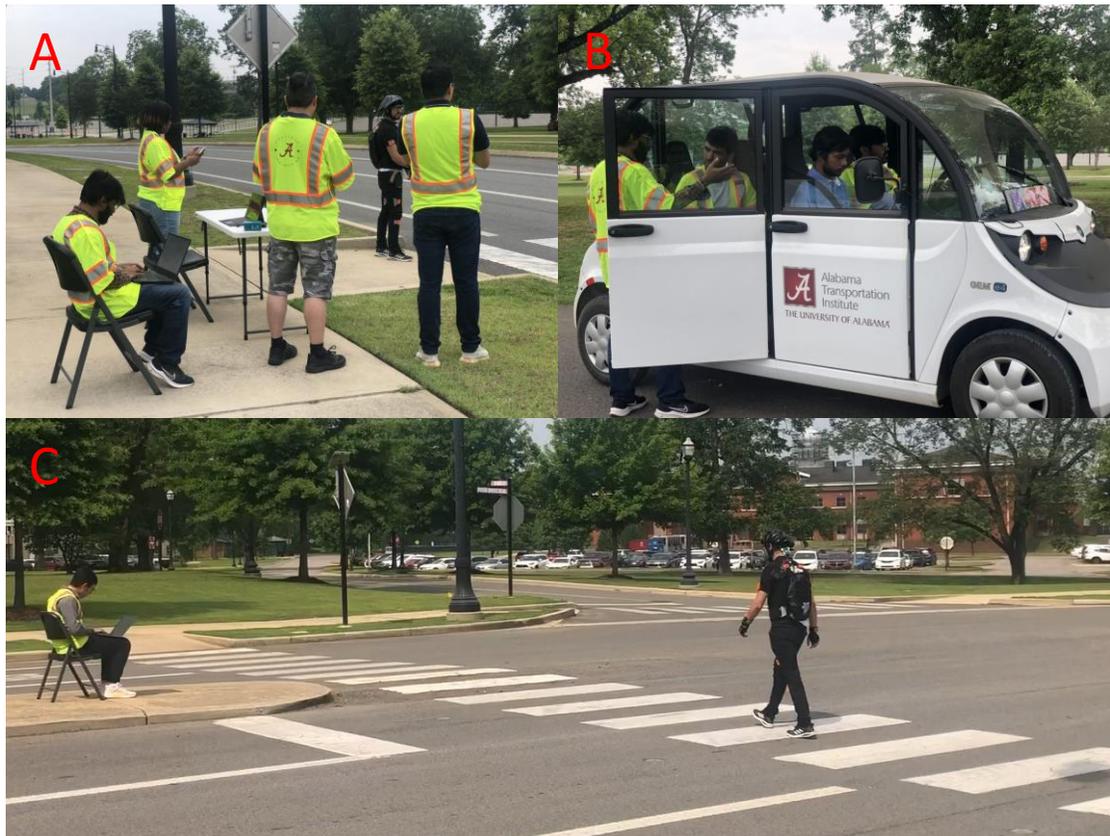

**FIGURE 3 A) Roadside computational unit; B) Sensors installed in the vehicle; and c) Pedestrian crossing the crosswalk with sensors installed in their backpack;**

*Communication gateway*

The sensors utilized for both the vehicle and the pedestrian are connected to their respective laptop computers via USB cable connections. Additionally, both laptops are connected to a Wi-Fi network, enabling the real-time uploading of collected data to a cloud-based database. The laptop installed in the vehicle serves as the processing unit for the DT, accessing pedestrian data in the cloud to execute the necessary computational tasks.

*Digital Twin*

*Digital shadow*: The digital shadow serves as a replica of the physical world, serving two main purposes: data processing and visualization on a map. The collected data are subjected to a structured definition, in which the variables of interest are carefully chosen and specified. In addition, indicrected variables are calculated from measurements of other variables. This is the case of distance traveled, which is used as one of the inputs in LSTM, and is calculated using the Haversine formula (22). Once all the desired data transformations are defined, they are ready to be applied to the data collected in real time. Regarding data visualization, trajectory matching techniques are applied to align the actual trajectories with predefined trajectories, ensuring accurate mapping and alignment of the data.

*Digital sibling:* This entire part of the Digital sibling is schematized in the **FIGURE 5**. In the digital sibling layer, the initial step involves predicting the distances traveled by both the pedestrian and the vehicle. This prediction task is accomplished using an LSTM recurrent neural network with an encoder-decoder architecture. As described in the scenario section, the vehicle has two possible movements: proceeding through or making a left-turn. Consequently, it is necessary to train and validate three separate LSTM models: one for pedestrians, one for the left-turn movement of the vehicle, and another for when the vehicle proceeds through. For each neural network, training and validation were conducted using data collected specifically for each of the three cases. With the trained LSTM models, parallel evalautions of potential safety crical scenarions are run to determine whether the vehicle and pedestrian are likely to crash. As shown in **FIGURE 2**, there are boxes (vehicle and pedestrian detection zones) indicating regions where predictions will be made if the vehicle or pedestrian is within them. This is because making predictions outside these zones would be irrelevant, since either they have already passed the potential collision zone, or there is still a considerable distance to go before they reach it. Therefore, using real-time sensor data from both the pedestrian and the vehicle, the first step is to verify whether they are within the risk zone. If both are within the zone, the LSTM pedestrian model is used to predict the pedestrian's distance based on the data from the previous steps. However, in the case of the vehicle, since it is not known whether it will go through or left-turn, predictions must be made using both trained LSTM models.

From the real-time data, a trajectory matching of the predefined trajectories is performed, and the predicted distances are used to determine the pedestrian and vehicle positions at subsequent time steps. This process is carried out for both the pedestrian and the vehicle, whether turning left or going through. To determine if a collision occurs, an algorithm incorporating the collision risk estimated (CRE) proposed by (23) is utilized. CRE is calculated by dividing the typical stop distance by the distance between the pedestrian and the vehicle. The typical stop distance refers to a longitudinal distance, but it is also essential to consider a lateral distance. According to (24), a collision occurs when the pedestrian is within the width of the vehicle. Therefore, we define a collision risk region (CRR) as a circular sector with a radius equal to the typical stop distance (i.e., CRR stopping distance) and a chord equal to twice the width of the vehicle (i.e., CRR width). This is done to prioritize safety since the sensor does not necessarily have to be in the center of the vehicle; it could be positioned on either side. For the 25-mph speed limit, CRR stoping distance is set to 16.95 meters (24). For the CRR width, which is twice the width of the vehicle, we will consider the maximum width of a vehicle as 2.6 meters (25). Using this CRR width, we calculate the angle alpha, which equals 8.72 degrees and will be used for further calculations. Alpha represents half of the angle of the circular sector. **FIGURE 4** illustrates a schematic of this collision risk region. Therefore, the calculation of CRR stopping distance and width (i.e., angle) and the calculation of CRR are done simultaneously for the pedestrian with respect to the vehicle going through and with respect to the vehicle turning left for each time step. If the angle and distance values of CRR are below the defined thresholds, it means that the pedestrian is within the CRR, and the CRE is passed to the next application step.

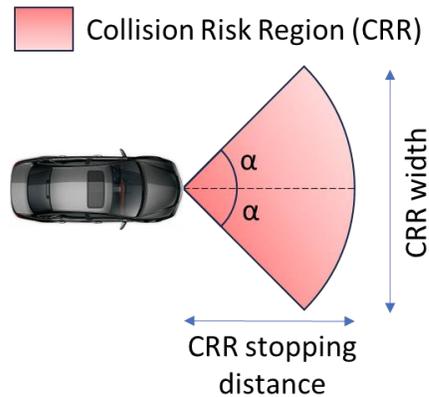

**FIGURE 4 Representaiton of collision risk region.**

*Application*: The objective of the application is to generate proactive safety alerts in case of a potential crash. This step immediately follows the digital twinning process (see **FIGURE 5**). Access to this step is granted only if the angle and distance values of CRR are below the defined threshold. Whenever CRE is above 1 there is a potential risk of crash. The further away from 1, the higher the risk. In such a case, a proactive VRU safety alerts are generated. In this case study, the proactive safety alerts are designed exclusively for the driver of the vehicle, in which the computer where all the DT is executed is located.

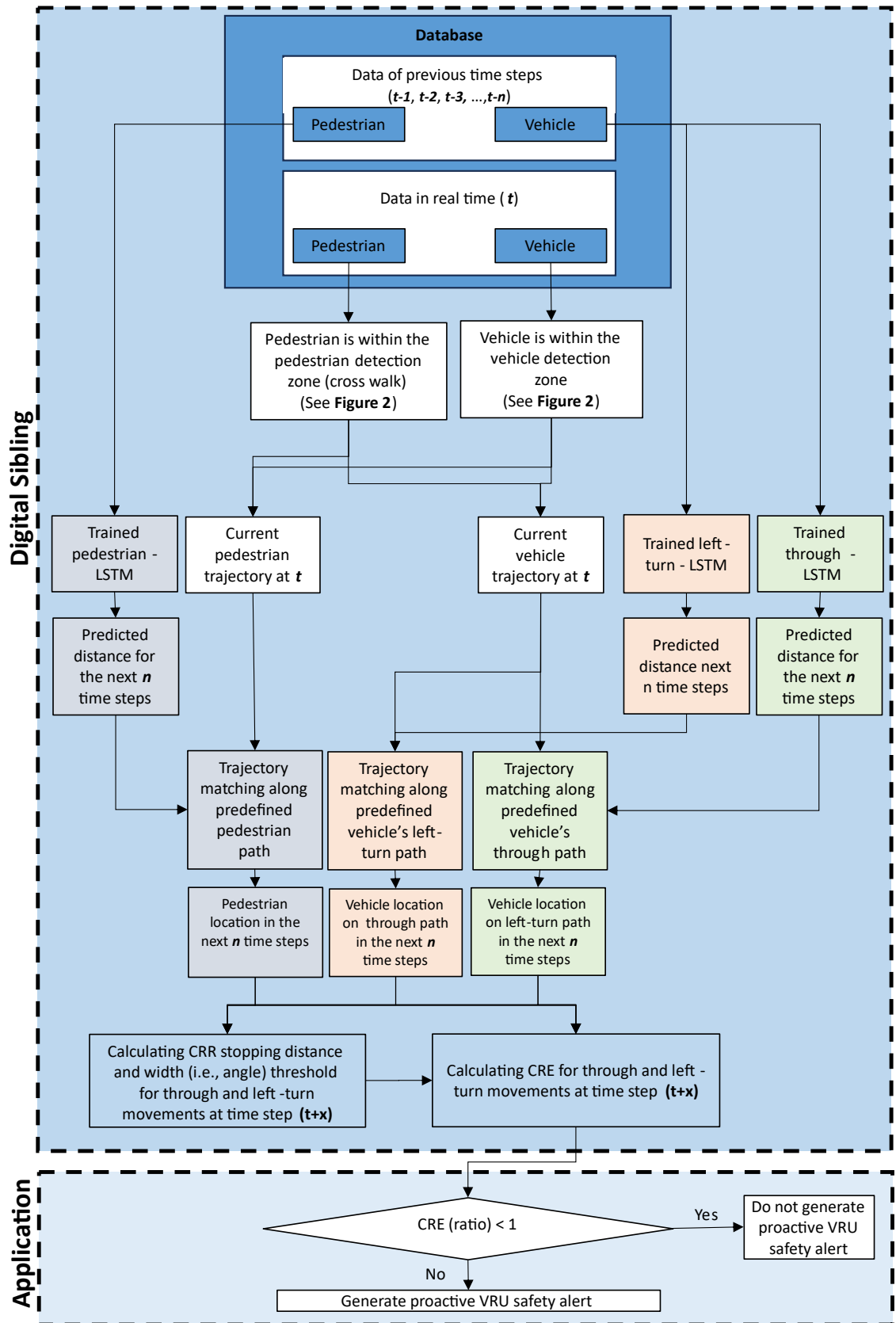

**FIGURE 5 Real-world implementation framework.**

**EVALUATION DT-BASED COLLISION WARNING SYSTEM**

This section will assess the results of the DT implementation. One of the key aspects of the DT is predicting the future positions of both pedestrians and vehicles, which is achieved through the LSTM model outcomes. Therefore, this section is divided into two parts. The first part evaluates the training and testing results of the three LSTM models: pedestrian, vehicle going through, and vehicle making a left-turn. The second part addresses the results of the complete DT implementation, showcasing the outcomes for the two potential collision scenarios: when the vehicle goes through and when the vehicle makes a left-turn.

**Evaluation of Pedestrian and Vehicle Trajectory Prediction**

Three LSTM networks have been trained and validated: pedestrian-LSTM, vehicle-through-LSTM, and vehicle-left-turn-LSTM. The architecture for all three cases follows an encoder-decoder LSTM model. The encoder consists of two layers, and the decoder comprises another two layers, with an additional dense layer used as the output for the network. Input and output data are organized into three-dimensional tensors with a shape of [samples, time steps, features]. The input time steps refer to past time steps used for prediction, while the number of time steps in the output corresponds to the prediction period. Features represent the input features used for prediction and the features being predicted in the output. The number of samples depends on the selected number of time steps. The data for each LSTM model has the same time difference between consecutive timestamps, which depends on the frequency of data generated by the pedestrian and vehicle sensors. The LSTM architecture is illustrated in **FIGURE 6**, and the input and output tensors are described in the **TABLE 2.**

The Encoder-Decoder LSTM is implemented using the Keras deep learning library in Python. The LSTM network is trained and validated by dividing the data into two sets and subsequently normalizing it within a range of 0 to 1. The number of neurons, epochs, and batch size are selected through a trial-and-error approach. These parameter values are listed in the **TABLE 2.** After testing each LSTM network, the Root Mean Square Error (RMSE) of the predicted distance is evaluated. These values are displayed in the **TABLE 2** for each of the three LSTM networks. **FIGURE 7** illustrates the mean absolute error (MAE) loss profile (learning curve), where the y-axis represents the mean absolute error loss for the training and validation datasets, and the x-axis represents the number of epochs.

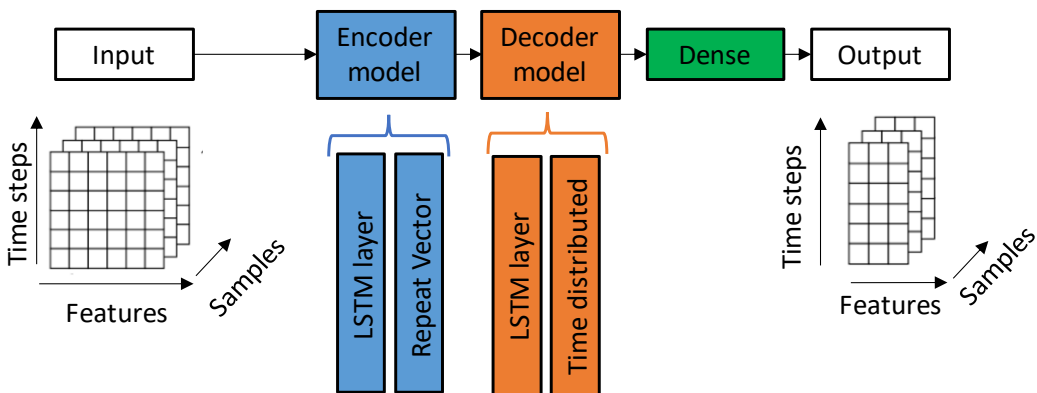

**FIGURE 6 Encoder-Decoder LSTM architecture along with the input and output tensors**

**TABLE 2 Summary of Model Parameters of the Trajectory Prediction Models**

|  | Pedestrian LSTM | Vehicle through LSTM | Vehicle left-turn LSTM |
|---|---|---|---|
| **Features Input** | Speed, Acceleraion (x,y,z), gyroscope (x,y,z), distance | | |
| **Time steps Input** | 4 | 10 | 10 |
| **Features Output** | distance | | |
| **Time steps Output** | 8 | 8 | 8 |
| **Number of neurons in Layer LSTM** | 128 | | |
| **Number of neurons in Repeat Vector** | 128 | | |
| **Number of neurons in Layer LSTM** | 64 | | |
| **Number of neurons in Layer Time distributed** | 1 | | |
| **Activation Function** | Rectified Linear Unit | | |
| **Optimizer** | Adam | | |
| **Learning rate** | 0.01 | | |
| **Root Mean Square Error (RMSE) (meters)** | 0.049 | 1.175 | 0.355 |

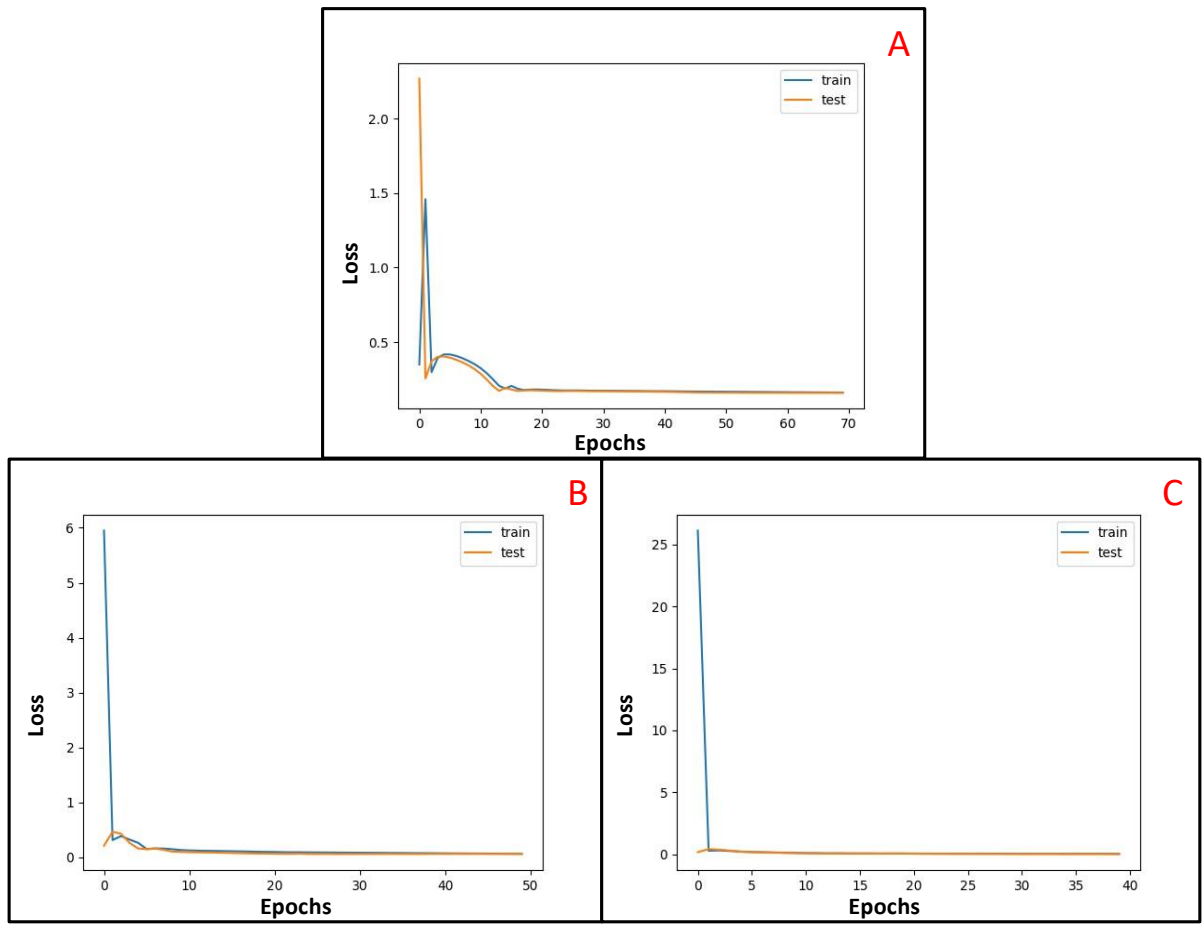

**FIGURE 7 Training and testing A) Pedestrian LSTM; B) Vehicle through LSTM; C) Vehicle left-turn LSTM.**

The three trained LSTM models exhibit similar learning curves with consistent behavior. In the three cases the learning curve demonstrates that the training loss decreases and stabilizes, indicating that the model is not underfitted. Moreover, the validation loss also decreases and stabilizes, showing that it is not overfitted. Additionally, the small gap between the validation and training losses indicates a good fit for the models. The RMS) for the pedestrian LSTM model is 0.049 meters. This value is low, considering that an error of approximately 5 centimeters, which is acceptable (as it would be within crosswalk width) for each second of pedestrian travel. Similarly, the RMSE for the vehicle left-turn LSTM model is 0.355 meters, which is also considered acceptable considering the width of a lane. However, for the LSTM predicting distances when the vehicle continues through, the error increases to 1.175 meters. This value becomes significant, indicating a potential error of over one meter for each second of vehicle travel.

**DT-enabled Proactive Safety Alert System Evaluation**

In this section, we proceed with the evaluation of the integrated LSTM models within the complete DT. As shown in **FIGURE 3**, data for both the pedestrian and the vehicle are collected through a real-world experiment. However, it was ensured that the pedestrian was never exposed to any potential collision or minimal risk during data collection. Consequently, no instances were found where the pedestrian and the vehicle simultaneously occupied the risk area at the same timestamp. However, we have data at different timestamps, allowing us to establish our baseline evaluation based on spatial proximity without considering the timestamps of the data. Given the aforementioned assumptions, the evaluation of the DT involves identifying potential collisions when pedestrians are situated within the CRR, as defined by the established thresholds without strict time alignment with the vehicle. Through this approach, a total of 10 potential collision points were detected, comprising 5 instances each for left-turn and through vehicle movements. These collision points form the foundation of our baseline evaluation, as illustrated in the **FIGURE 8.**

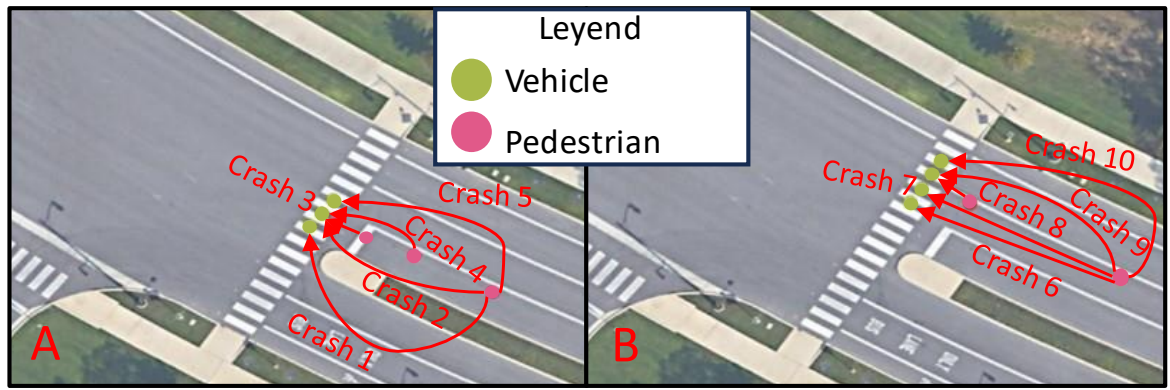

**FIGURE 8 Potential baseline crash scenarios independednt of time stamp: A) Left-turn; B) Through**

With the established baseline, we validated our DT by forcing the coincidence of the timestamps of the 10 baseline crashes. The matched crashes are 4, which are crashes 1 and 2 for the left-turn and crashes 9 and 10 when the vehicle goes through. Notably, crashes 1 leads to both the vehicle and the pedestrian moving forward for 1 second before encountering collision 2. For our safety-critical scenarios, we assume that crash 1 occurs at the predicted time *t+3*, and crash 2 at *t+4*. **FIGURE 9** illustrates both potential crashes schematically taking into accounte the timestamps matched. For the crashes 9 and 10 where the vehicle makes a left-turn, we consider that the crashes occur at a predicted time *t+7* and *t+8*. This evaluation allows to assess the performance of the DT in longer-term predictions, as these instances are expected to have higher prediction errors.

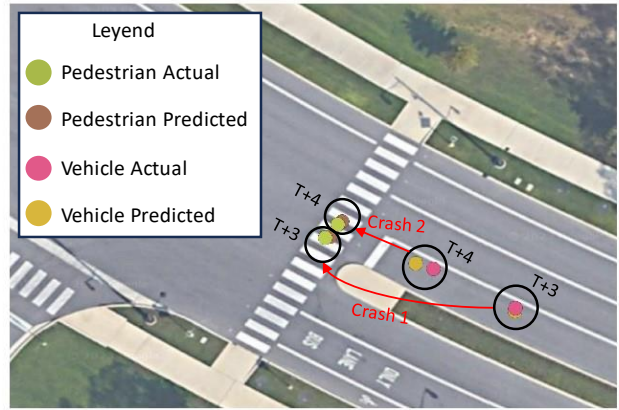

**FIGURE 9 Crash 1 and 2 detedted by DT in time t+3 and t+4 respectively**

**TABLE 3** presents the results of the DT evaluation for these 4 considered crashes. As mentioned earlier, the CRR stopping distance and angle, as well as the CRE, are calculated simultaneously. **TABLE 3** provides the ground truth values and the predicted values for CRR. As observed, all values are below the defined thresholds of 8.72 degrees for the angle and 16.95 meters for the distance. The CRE is also calculated for both ground truth and prediction, as shown in the **TABLE 3**. For the CRE calculation, the typical stop distance defined as 16.95 meters is divided by the calculated distance from the pedestrian to the vehicle. Notably, the CRE values for all 4 accidents are greater than 1 for both ground truth and prediction. This indicates that in all considered potential crasehs, there is a risk of collision, and the DT correctly detects them. These findings indicate that the DT performs adequately even for predictions further ahead in time, supporting its reliability and effectiveness in potential safety alerts.

**TABLE 3 Calculation of CRR and CRE for Ground Truth and Predicted**

| Potential future crash scenarios in which the vehicle and the pedestrian time stamps matched each other | CRR stopping distance and width (i.e., angle) Treshold angel = 8.72 degrees Treshold distance = 16.95 meters | | | | CRE = Typical stop distance divied by pedestrian-vehicle distance. | | | | | | Evaluation (Treshold is 1) |
|---|---|---|---|---|---|---|---|---|---|---|---|
| | Ground truth | | Predicted | | Ground truth | | | Predicted | | | |
| | Distance (meters) | Angle (degrees) | Distance (meters) | Angle (degrees) | Typical stop distance (meters) | Pedestrian-vehicle distance. (meters) | CRE | Typical stop distance (meters) | Pedestrian-vehicle distance. (meters) | CRE | |
| **Crash 1 (time t+3)** | 14.57 | -5.32 | 14.38 | -8.46 | 16.95 | 14.57 | 1.16 | 16.95 | 14.38 | 1.18 | Crash |
| **Crash 2 (time t+4)** | 7.66 | -1.33 | 6.24 | 0.58 | 16.95 | 7.66 | 2.21 | 16.95 | 6.24 | 2.72 | Crash |
| **Crash 9 (time t+8)** | 16.59 | 6.69 | 11.38 | 3.28 | 16.95 | 16.59 | 1.02 | 16.95 | 11.38 | 1.49 | Crash |
| **Crash 10 (time t+7)** | 16.52 | 2.51 | 10.58 | -4.09 | 16.95 | 16.52 | 1.03 | 16.95 | 10.58 | 1.60 | Crash |

Regarding the DT results, several combinations can be used to generate potential collisions by modifying the timestamps. The chosen four accidents cover a range from a prediction time of 3 seconds to the maximum possible of 8 seconds. Despite a variations in distance and angle, the overall results are promising. These differences can be attributed to the LSTM models trained. Improving this LSTM model by training it with more data is likely to yield better results.

## CONCLUSIONS

This paper introduces a TDT approach designed specifically for pedestrian-vehicle interactions, determining the future potential collision scenarios and generating proactive safety alerts. The objective of this study is to utilize Digital Twin (DT) technology to enable a proactive safety alert system for VRUs. A pedestrian-vehicle trajectory prediction model has been developed using the Encoder-Decoder Long Short-Term Memory (LSTM) architecture to predict future trajectories of pedestrians and vehicles. Subsequently, parallel evaluation of all potential future safety-critical scenarios is carried out. Three Encoder-Decoder LSTM models, namely pedestrian-LSTM, vehicle-through-LSTM, and vehicle-left-turn-LSTM, are trained and validated using field-collected data, achieving corresponding root mean square errors (RMSE) of 0.049, 1.175, and 0.355 meters, respectively. A real-world case study has been conducted where a pedestrian crosses a road, and vehicles have the option to proceed through or left-turn, to evaluate the efficacy of DT-enabled proactive safety alert systems. Experimental results confirm that DT-enabled safety alert systems were succesfully able to detect potential crashes and proactively generate safety alerts to reduce potential crash risk. In conclusion, the proposed approach shows promising preliminary results, which can be enhanced with more extensive data. In this work, we focus on pedestrian and human-driven connected vehicles. However, this research could be extended to a broader context that also includes the coexistence of Connected and Autonomous Vehicles (CAVs). Interactions in this mixed traffic environment will be difficult due to the complexities arising from the automated nature of CAVs, as their automated behavior can lead to misunderstandings with humans. In this way, public trust and perception towards AVs can be improved, as accidents involving AVs have a greater impact and tend to create negative public opinion about them.

## ACKNOWLEDGMENTS

This material is based on a study supported by Alabama Transportation Institute (ATI) and Alabama Transportation Policy Research Center (TPRC). This work has also been funded by the Ministry of Science and Innovation of the Government of Spain through the Predoctoral Contract Grant (PCG) with reference PRE2020-096222. Any opinions, findings, and conclusions or recommendations expressed in this material are those of the author(s) and do not necessarily reflect the views of the ATI, TPRC and PCG, and the ATI and TPRC assume no liability for the contents or use thereof.

## AUTHOR CONTRIBUTIONS

The authors confirm their contribution to the paper as follows: study conception and design: E. Rúa, K. Shakib, S. Dasgupta, M. Rahman, and S. Jones; data collection: E. Rúa, K. Shakib, S. Dasgupta; interpretation of results: E. Rúa, K. Shakib, S. Dasgupta, M. Rahman, and S. Jones; draft manuscript preparation: E. Rúa, K. Shakib, S. Dasgupta, M. Rahman, and S. Jones. All authors reviewed the results and approved the final version of the manuscript.